\documentclass[letterpaper, 10 pt, conference]{ieeeconf}  

\IEEEoverridecommandlockouts                              

\usepackage{graphics} 
\usepackage{epsfig} 
\usepackage{mathptmx} 
\usepackage{times} 
\usepackage{amsmath} 
\usepackage{amssymb}  
\usepackage{subcaption}
\usepackage{multirow}
\usepackage{listings}
\usepackage{xcolor}
\usepackage{booktabs}
\usepackage{svg}
\usepackage{lmodern}
\usepackage{graphicx}
\usepackage{balance}
\graphicspath{{image/}} 

\title{\LARGE \bf
ASGrasp: Generalizable Transparent Object Reconstruction and 6-DoF Grasp Detection from RGB-D Active Stereo Camera}

\author{Jun Shi$^{1}$, Yong A$^{1}$, Yixiang Jin$^{1}$, Dingzhe Li$^{1}$, Haoyu Niu$^{1}$, Zhezhu Jin$^{1}$, He Wang$^{2,3,4\dag}$
\thanks{$^{1}$Samsung R\&D Institute China-Beijing}
\thanks{$^{2}$CFCS, School of Computer Science, Peking University}%
\thanks{$^{3}$Galbot}%
\thanks{$^{4}$Beijing Academy of Artificial Intelligence (BAAI)}%
\thanks{\dag Corresponding to hewang@pku.edu.cn.}%
}

\begin{document}

\maketitle
\thispagestyle{empty}
\pagestyle{empty}

\begin{abstract}
In this paper, we tackle the problem of grasping transparent and specular objects.  
This issue holds importance, yet it remains unsolved within the field of robotics due to failure of recover their accurate geometry by depth cameras.
For the first time, we propose ASGrasp, a 6-DoF grasp detection network that uses an RGB-D active stereo camera. ASGrasp utilizes a two-layer learning-based stereo network for the purpose of transparent object reconstruction, enabling material-agnostic object grasping in cluttered environments.
In contrast to existing RGB-D based grasp detection methods, which heavily depend on depth restoration networks and the quality of depth maps generated by depth cameras, our system distinguishes itself by its ability to directly utilize raw IR and RGB images for transparent object geometry reconstruction.
We create an extensive synthetic dataset through domain randomization, which is based on GraspNet-1Billion. 
Our experiments demonstrate that ASGrasp can achieve over 90\% success rate for generalizable transparent object grasping in both simulation and the real via seamless sim-to-real transfer. Our method significantly outperforms SOTA networks and even surpasses the performance upper bound set by perfect visible point cloud inputs. Project page: \textcolor{magenta}{https://pku-epic.github.io/ASGrasp}
\end{abstract}

\section{INTRODUCTION}
Recent years have witnessed enormous progress \cite{fang2020graspnet, breyer2021volumetric, jiang2021synergies, sundermeyer2021contact, gou2021rgb, wang2021graspness} in the field of learning-based diffuse object grasping from depth observations. However, commercial depth sensors fail to accurately sense transparent and specular objects, therefore grasping these kind of objects has become a major bottleneck in developing full 3D sensor solution to general grasping tasks.  



Several works \cite{dai2022domain, fang2022transcg, sajjan2020clear} have tackle this challenging problem by learning to estimate or restore the depth of the transparent and specular objects. 
ClearGrasp\cite{sajjan2020clear} directly removes the transparent area from the raw depth map and then uses an optimization technique to restore the depth. We note that this method fails to fully utilize the original depth observation and relies on precise segmentation of the transparent objects. 
Later on, DREDS\cite{dai2022domain} and TransCG\cite{fang2022transcg} propose to learn a mapping from the raw depth along with its RGB observation to ground-truth depth. However, the
correct depth information of the transparent objects may already be gone in the original depth, due to errors or failures in stereo matching, and therefore taking the complete raw depth as input wouldn't help much, either.
Recently, GraspNeRF~\cite{dai2023graspnerf} and EvoNeRF~\cite{kerr2022evo} get rid of depth sensors and instead reconstruct the shape of transparent objects from multi-view RGB images, at the cost of extra time cost for multi-view image capture on a mobile robot.   
Overall, none of these methods have gained informative cues of the transparent area from the depth observations, thus leaving a huge space to improve further reconstruction and grasping of the transparent objects.

In this work, we propose to leverage the raw IR observations from an active stereo camera that are commonly found in commercial products(e.g., Intel R200 and D400 family~\cite{zhang2018activestereonet}) to improve transparent object depth estimation and grasping detection. Our insight is that the left and right IR observations, before going through error-prone stereo matching, carry the original information of transparent object depth while RGB information can provide extra shape priors. We thus devise an RGB-aware learning-based stereo matching network that take inputs both RGB and two IR images, based upon a popular flow estimation network\cite{lipson2021raft}. 



\begin{figure}[t]
    \centering
    \includegraphics[width=\linewidth]{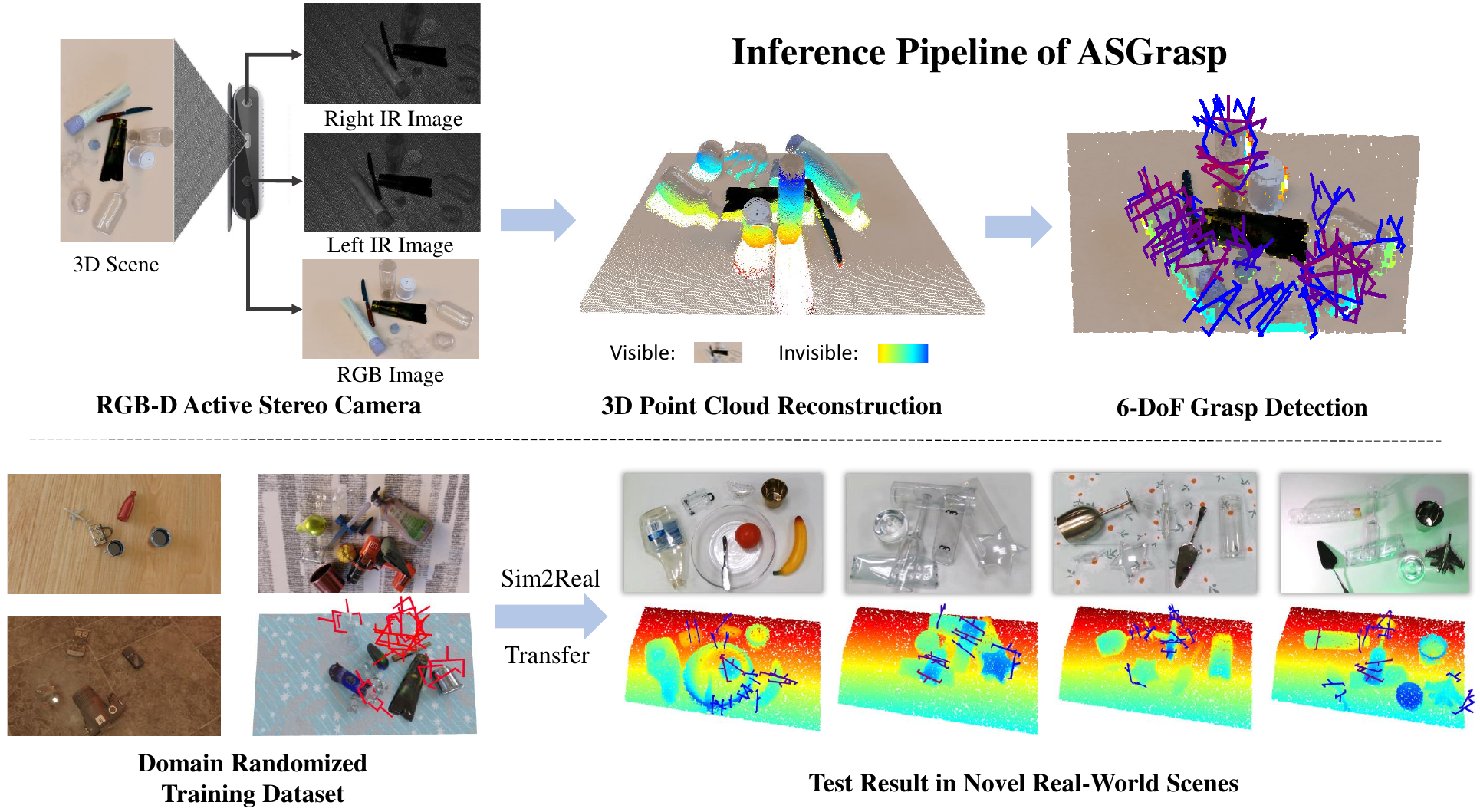}
    \setlength{\abovecaptionskip}{-5pt}
    \setlength{\belowcaptionskip}{-15pt}
    \caption{Overview of proposed method ASGrasp and the dataset. Our approach takes one RGB image and left-right infrared (IR) pair as inputs, predicts visible and invisible point cloud to generate grasp pose. We train the model using the DREDS dataset and the rendered dataset based on GraspNet-1Billion  in complex scenes. }
    \label{fig:1}
\end{figure}
Our experiments show that the predicted depth is of very high quality and passing its backprojected point cloud to the SOTA point cloud based grasping detection network, GSNet\cite{wang2021graspness} can already be on par with using perfect depth point cloud input.
We suspect that, limited by the single-view nature of the depth point cloud input, GSNet may still lack sufficient geometric cue of the invisible region, especially when lateral grasping is needed. To surpass the performance of perfect depth point cloud, we further propose to predict the second-layer depth to approximate the invisible point cloud, inspired by \cite{shin20193d}. 
Via combining the visible and invisible predicted point clouds and using them as inputs to GSNet, the grasping performance even outperforms when taking input the perfect depth point cloud.

To train our networks, we extend the GraspNet-1Billion\cite{fang2020graspnet} dataset to a large-scale photorealistic synthetic dataset of objects with diffuse, transparent, or specular material. Comprising 115,000 sets of RGB and IR images, this dataset is generated through physics-based rendering and enhanced with diverse domain randomization. Our network exhibits the ability to readily generalize to novel real-world scenes after trained on this extensive dataset.


In summary, our main contributions are as follows:
\begin{enumerate}

\item We propose a novel RGB-aware two-layer stereo network for generalizable transparent object reconstruction from RGB-D active stereo camera.
\item To the best of our knowledge, we, for the first time, achieve over 90\% success rate for generalizable transparent object grasping in both simulation and the real, without seeing any real training data.

\end{enumerate}

\section{RELATED WORK}
\subsection{Active Stereo Based Depth Estimation}
Active stereo camera projects a texture into the scene using an IR projector, effectively addressing the depth estimation challenge in textureless scenes when compared to passive stereo sensors\cite{konolige2010projected}. However, it still grapples with common stereo matching issues. ActiveStereoNet\cite{zhang2018activestereonet} is a fully self-supervised method that produces precise depth with subpixel precision, preserves edges, handles occlusions, and avoids over-smoothing issues. ActiveZero\cite{liu2022activezero} combining supervised disparity loss and self-supervised losses to train active stereo vision systems without the need for real-world depth annotation. Active stereo cameras such as Intel RealSense include stereo cameras and RGB cameras\cite{keselman2017intel}. Following the MVS framework, a triview stereo system was created in this paper by combining RGB with left IR image and right IR image, aiming to explore its effectiveness in transparent and specular scenes.

\subsection{Transparent And Specular Scene Depth Completion}
Current commercial depth sensors fail to capture depth images for transparent and specular objects.
 ClearGrasp\cite{sajjan2020clear} makes key modifications to the two-stage depth completion pipline, trying to optimize the depth image by using RGB.
 DepthGrasp\cite{tang2021depthgrasp}  modifies the global optimization, which partially improved the performance. \cite{zhu2021rgb} uses the local implicit function for depth completion of transparent objects. DREDS\cite{dai2022domain} and TransCG\cite{fang2022transcg} attempt to restore depth utilizing RGB information and can be considered as a one-stage depth completion. Compared to the above depth completion methods, the two-layer depth estimation proposed in this paper can complete invisible depth from the current viewpoint.

\subsection{Grasp Network}
Learning-based approaches play a significant role in the field of robot grasping\cite{bohg2013data}, generating 6 DoF grasps from a point cloud or TSDF\cite{breyer2021volumetric}. GSNet\cite{wang2021graspness} takes a dense scene point cloud as input and uses a graspness-based sampling strategy to select points with high graspness.  
AnyGrasp\cite{fang2023anygrasp} generates grasp poses from a partial point cloud using a geometry processing module.
VGN\cite{breyer2021volumetric} directly outputs the predicted grasp quality, gripper orientation, and opening width for each voxel in the queried TSDF volume.
GIGA\cite{jiang2021synergies} takes the input TSDF volume and generates grasp proposals based on the learned implicit functions. Point cloud can represent irregular and non-uniformly sampled data, making them suitable for complex, unstructured environments. In this paper, we choose GSNet as grasp network.

\section{METHOD}
\subsection{Problem Statement and Method Overview}
In the context of an active stereo system comprising an RGB camera and two infrared (IR) cameras capturing a single viewpoint of a cluttered tabletop scene containing transparent, specular, and diffuse objects, the objective of the proposed robotic system is to detect 6-DoF grasping poses and subsequently execute grasping-to-removal operations for all the objects.
The intrinsic and extrinsic parameters of the active stereo system are acquired from the factory settings.
Next, we define the 6D grasp detection learning task as the mapping of a set of one RGB image $I_c\in R^{H\times W\times 3}$ and two IR images $I_{ir}^{l},I_{ir}^{r}\in R^{H\times W}$ to a set of 6-DoF grasp poses $\{g_j|g_j=(q_j,t_j,R_j,\omega_j)\}$. Each detected grasp pose includes grasp score $q_j\in [0,1]$, grasp center $t_j\in R^3$, rotation$R_j\in SO(3)$, and opening width $\omega_j$.

Our proposed framework consists of two main components: the scene reconstruction module, denoted as $F_{d}$, and the grasp detection module based on explicit point cloud, denoted as $F_g$, as illustrated in the Fig. 2. In $F_{d}$, we choose the RGB image $I_c$ as the reference image and differentiably warp the left $I_{ir}^{l}$ and right IR images $I_{ir}^{r}$ to the RGB reference coordinate system, constructing a cost volume $C\in R^{H\times W\times D_h}$. This allows us to leverage the classical GRU-based stereo matching method\cite{lipson2021raft}. Additionally, we introduce a second-layer depth branch that enables the network to predict not only the visible depth, refer to it as the first-layer depth $D_{1}\in R^{H\times W}$ but also attempts to recover second-layer depth $D_{2}$ for objects in the scene, which help to capture the complete 3D shape of unoccluded obejcts. In $F_g$, we employ the two-stage grasping network $F_{gsnet}$\cite{wang2021graspness}, and thanks to the richer input point cloud information, the $F_{gsnet}$ can predict more accurate grasp poses in the second stage.

\subsection{Scene Reconstruction}
For single-view grasping networks, one crucial aspect is the accurate recovery of scene depth, especially for transparent and specular objects. Additionally, the single-view depth only record the first intersection point of rays with the surfaces of objects in the scene. With only partial object shape information available, predicting optimal grasp poses becomes a more challenging task for the grasping network, particularly when dealing with lateral grasping orientations. To address this challenge, we propose a framework based on RAFT-like stereo matching that simultaneously recovers first layer depth for transparent and specular objects as well as occluded surface information for objects by second-layer depth prediction.

\textbf{Image Features}  Following \cite{lipson2021raft}, given the left and right IR images $I_{ir}^{l},I_{ir}^{r}$ and RGB image $I_c$, we use a feature encoder to extract IR image features ${F}_{ir}^{l}, {F}_{ir}^{r}$ at 1/4 of the  $I_{ir}^{l(r)}$ image resolution, and a context encoder to extract multi-scale context features ${F}_c$ at 1/4,1/8,1/16 of the RGB image resolution with 128 channels.

\textbf{Epipolar Cost Volume} Following \cite{ma2022multiview,wang2022itermvs}, after extacting feature maps from IR image pairs, we firstly use differentiable bilinear sampling to warp the IR features w.r.t reference RGB view at the given depth hypotheses and then construct a 3D cost volume by computing the correlation between them. Specifically, with known camera intrinsics $\left\{{K_{c}, K_{ir}}\right\}$ and relative transformations$\left\{\left[{R}_{c\rightarrow j}|{t}_{c\rightarrow j} \right] \right\}$ between RGB and IR cameras ($j=l/r$ represents left or right IR, respectively), for each pixel ${p}_c$ in the RGB view,  and the depth hypothesis $d:=d({p})$, we can compute the corresponding pixel features in IR images as:
\begin{equation}
    {p}_{j} = {K}_{j}\cdot ({R}_{c\rightarrow j}\cdot({K}_c^{-1} \cdot {p}_c \cdot d) + {t}_{c\rightarrow j})
\end{equation}
and the cost volume $C\in R^{H\times W\times D_{h}}$can be computed as ${c}({p}) = \left \langle {F}_{ir}^{l}({p}_{l}), {F}_{ir}^{r}({p}_{r})\right \rangle$,
where $\left \langle \cdot, \cdot \right \rangle$ denotes the dot product, and $D_{h}$ is the number of the depth hypotheses. As all depths hypotheses sampled uniformly over the inverse depth range, we proceed to represent the iterative outputs in the disparity field. 

\textbf{Iterative Update } Besides original first-layer disparity $\mathit{d}_1$, we introduce an additional branch to update a second-layer disparity $\mathit{d}_2$ to predict invisible shapes of objects. Both $\mathit{d}_1$ and $\mathit{d}_2$ are initialized to zero. In each iteration $i$, per-pixel first and second-layer cost features are extracted from the cost volumes using the current disparity $\mathit{d}_1^{(i)},\mathit{d}_2^{(i)}$ respectively. These features include a hidden state $h^{(i)}$, the context features $\mathit{F}_c^{(i)}$ from the RGB image, which are inputted to the GRU-based update operator. The update operator outputs all increments $\mathit{\Delta}{d}_{k}$ of the first and second disparity fields as well as a new hidden state. Then we update both the first and second-layer disparities as $ \mathit{\Delta} {d}_{k+1} = \mathit{d}_{k}+\mathit{\Delta}{d}_{k}$,
where $\mathit{k}$ denotes the number of iterations.
\subsection{Point Cloud Based Grasp Detection}
Our point cloud based grasp detection network is built upon the SOTA work, GSNet~\cite{wang2021graspness}.
The original GSNet is a two-stage grasp detector to predict dense grasp poses from a single-view point cloud. 
To extract comprehensive geometric features of the scene, our network ${F}_g$ takes not only visible but also invisible point clouds as inputs, which are obtained by sampling from the first and second-layer depths, respectively. The visible point cloud serves as the primary resource for generating augmented grasp points for the subsequent neural network, while the invisible point cloud is used as a reference to provide additional contextual features.

\begin{figure*}[htp]
\centering
\includegraphics[width=\linewidth]{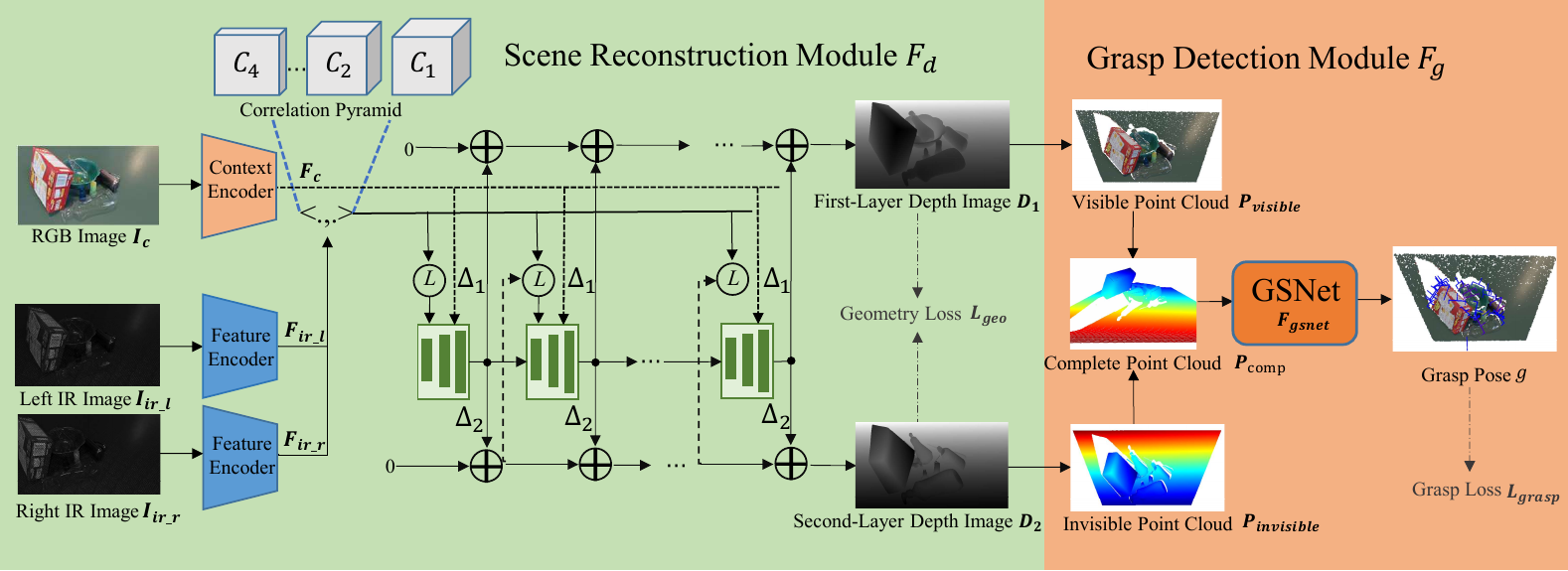}
\setlength{\abovecaptionskip}{-5pt}
\setlength{\belowcaptionskip}{-15pt}
 \caption{The architecture of our proposed approach involves the extraction of features $\protect{F_c}$, $\protect{F_{ir}^{l}}$, and $\protect{F_{ir}^{r}}$ from the RGB image $\protect{I_c}$, the left IR image $\protect{I_{ir}^{l}}$, and the right IR image $\protect{I_{ir}^{r}}$. $\protect{F_{ir}^{l}}$ and $\protect{F_{ir}^{r}}$ are used to construct a correlation pyramid. This correlation pyramid, along with $\protect{F_c}$, is fed into a GRU network for the prediction of a complete point cloud, which includes the first-layer depth (visible point cloud) and the second-layer depth (invisible point cloud). Subsequently, GSNet generates grasp poses based on the complete point cloud.}
\label{fig:2}
\end{figure*}

\subsection{Synthetic Dataset Generation}
Existing grasping datasets like \cite{fang2020graspnet} don't include transparent and specular objects, while current datasets for these materials lack dense grasping annotations\cite{fang2022transcg, sajjan2020clear, zhu2021rgb, liu2020keypose}.
We introduce a synthetic grasping dataset featuring transparent and specular objects which comprises 115k sets of RGB and IR images, along with 1 billion grasp poses.

We create this dataset using a data generation pipeline based on \cite{dai2022domain}. 
Initially, we modify the object materials in \cite{fang2020graspnet} to include diffuse, specular, or transparent properties, so we name it STD-GraspNet. 
Subsequently, we use Blender\cite{blender2018blender} to generate photorealistic RGB and IR images, utilizing the settings of the Realsense D415 camera.
To address the challenge of sim-to-real gap, we apply domain randomization in \cite{dai2022domain}.
In terms of camera poses, we not only use the original poses from \cite{fang2020graspnet},   but also add 3k additional random camera poses to enhance diversity and mimic various distances to objects.
We train our networks with STD-GraspNet, treating real data as a training data variation, thus improving real-world performance.
\subsection{Network Training}
Our training objectives consist of two main components and we train them separately:

\textbf{Geometry Loss}
We follow \cite{lipson2021raft} to calculate the L1 loss on all predicted first and second-layer disparities $\{{d}_{1}^{(i)},{d}_{2}^{(i)}\}_{i=1}^{N}$. Same to \cite{dai2022domain}, we add higher weight to the loss with in the transparent and specular objects, to concentrate more on the depth completion:
\begin{equation}
L_{geo} = {\sum_{i=1}^{N}}{\gamma}^{N-i}(||d_{1}^{*}  - d_{1}^{(i)}||_{1} + ||d_{2}^{*}-d_{2}^{(i)}||_{1})  
\label{eq.1}
\end{equation}
where {$\gamma$}=0.9,  $d_{1}^{*}$  and  $d_{2}^{*}$  are respresented as ground truth disparties.

\textbf{Grasping Loss}
For grasping learning, we supervise the point-wise graspable landscape, view-wise graspable landscape, grasp scores and gripper widths as in GSNet\cite{wang2021graspness}. 
Although we predict a complete point cloud, we still calculate the loss based on the visible point cloud. The whole objective can be formulated as:
\begin{equation}
{L_{grasp} = L_{o} + {\alpha}(L_p+ {\lambda}L_v) + {\beta}(L_s + L_w)} 
\label{eq.2}
\end{equation}
where $L_{o}$ is for objectness classification, $L_p$, $L_v$, $L_s$ and $L_w$ are for regressions of point-wise graspable landscape, view-wise graspable landscape, grasp scores and gripper widths respectively. $L_p$ and $L_s$ are calculated when the related points are on objects, $L_v$ is calculated for views on seed points and $L_w$ is calculated for grasp poses with ground truth scores $>$ 0. We use softmax for classification task and smooth-L1 loss for regression tasks.

\section{EXPERIMENTS}
\subsection{Implementation Details}
\textbf{Scene reconstruction network}. We implement the  network with PyTorch. For all training, we use the AdamW optimizer and clip gradients to the range $[-1,1]$. On DREDS-CatKnown, we train the model with the  first-layer depth loss for 100k steps with a batch size of 4. On STD-GraspNet training split, we finetune the pre-trained model for another 100k steps. For all experiments, we use 12 update iterations during training.

\textbf{Grasp detection network}. The input point cloud is cropped with a depth range of [0.25m, 1.0m]. The visible point cloud is down-sampled to 15000 for training and 25000 for inference, and invisible point cloud is always sampled to 10000. The graspness threshold is 0 for both training and inference. The model is trained on four NVIDIA RTX 3090 GPU that takes about 1 day for 10 epochs with a batch size of 4. We utilize the Adam optimizer with an initial learning rate of 0.001 and decay 5\% every epoch.

\subsection{Experiment Setup}
\textbf{Hardware Setup}. We use a 7-DoF Franka Panda arm with its default two-finger gripper, on which we mount an active stereo RGB-D camera, Intel Realsense D415.

\textbf{Simulation Setup}. We use PyBullet~\cite{coumans2016pybullet} for physical simulation of grasping. We use an active stereo sensor simulation~\cite{dai2022domain} built upon Blender\cite{blender2018blender} to render realistic RGB and IR images.



\begin{figure}[t]
    \centering
    \includegraphics[width=\linewidth,height=0.28\textheight]{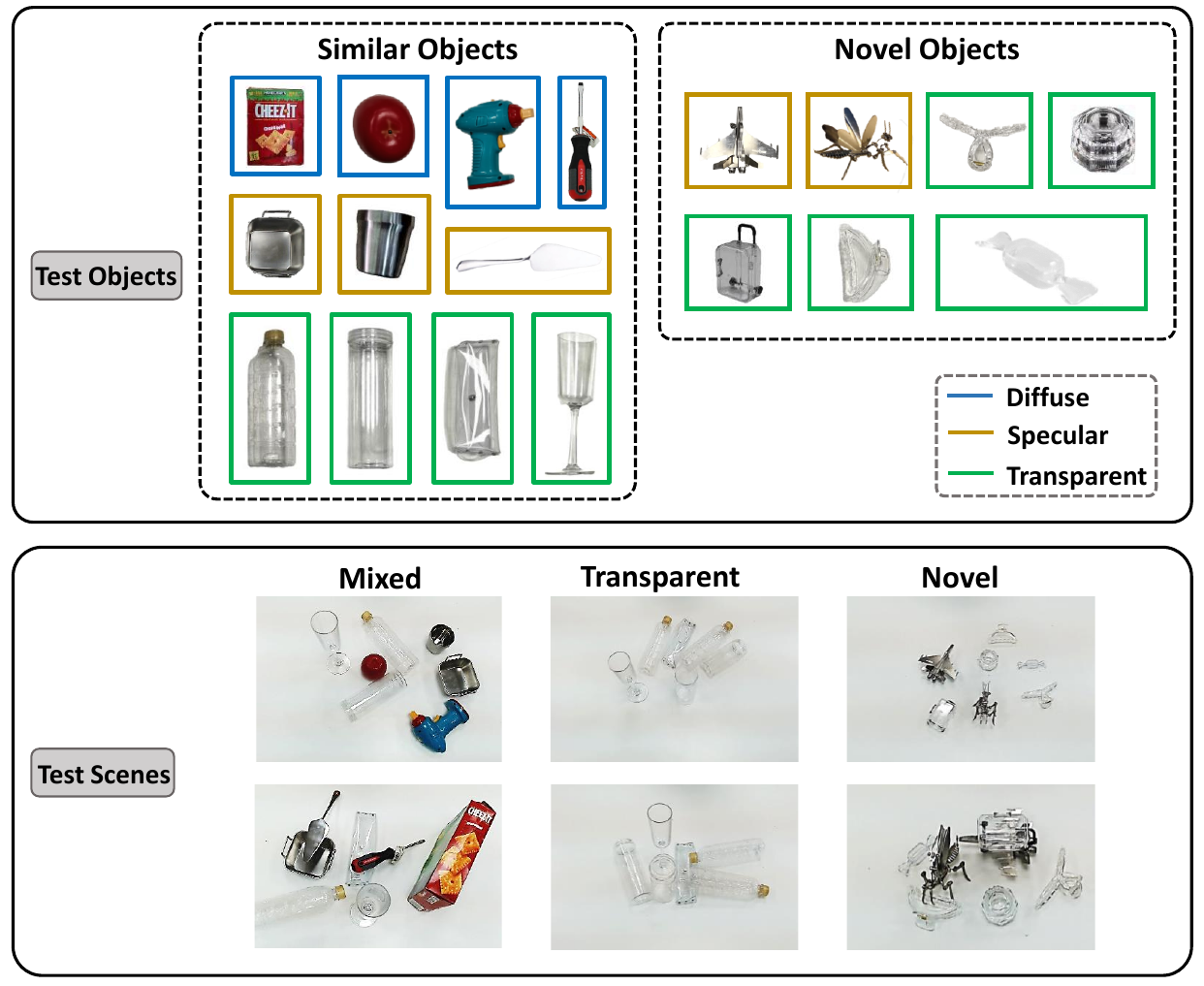}
    \setlength{\abovecaptionskip}{-5pt}
    \setlength{\belowcaptionskip}{-18pt}
    \caption{Real world evaluation dataset}
    \label{fig:real_world_setting}
\end{figure}
\subsection{Scene Reconstruction Experiments}
\textbf{Evaluation Metrics.} 
We evaluate the performance of transparent objects depth completion by three metrics, following in~\cite{zhu2021rgb}. 1) \textbf{RMSE}: the root mean squared error, 2) \textbf{REL}: the mean absolute relative difference, 3) \textbf{MAE}: the mean absolute error. For the first-layer depth, we resize the prediction and GT to the resolution of $126 \times 224$ for fair comparison, while for the second-layer depth, we resize them to $224\times224$ resolution consistent with the SwinDR. We evaluate all objects area and challenging area (specular and transparent objects), respectively.

\begin{figure*}[t]
    \centering
    \setlength{\abovecaptionskip}{-0pt} 
    \setlength{\belowcaptionskip}{-15pt}
    \includegraphics[width=\linewidth]{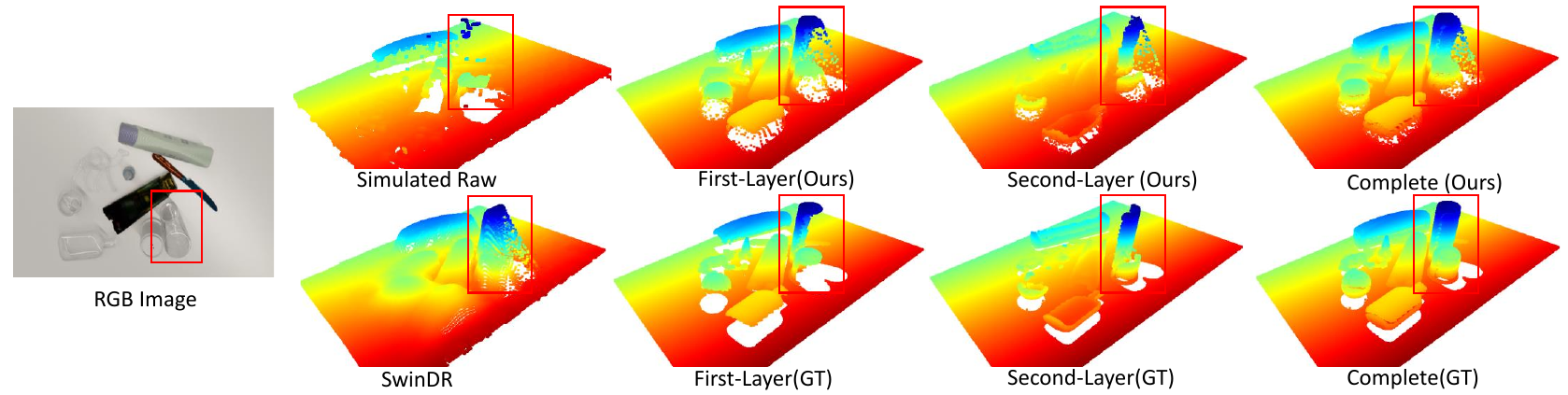}
    \caption{Qualitative comparison of point cloud reconstruction for an exemplar test data from STD-GraspNet. }
    \label{fig_qualit}
\end{figure*}

\textbf{Result comparisons.} We first compare our first-layer depth with SOTA methods on the test split of DREDS-CatKnown~\cite{dai2022domain}, including LIDF~\cite{zhu2021rgb}, NLSPN~\cite{park2020non} and SwinDR~\cite{dai2022domain}. All baselines are trained on the training split of DREDS-CatKnown. Subsequently, we compare our results with SwinDR on STD-GraspNet test split after fine-tuning on STD-GraspNet training split.
As shown in Table \ref{tab:dep_qua} and Table \ref{tab:dep_qua1}, we achieve the best performance compared to other methods on DREDS and STD-GraspNet for the first-layer depth.  Due to the absence of a suitable baseline, we can only report our performance for the second-layer depth. We also provide a qualitative comparison of the predictions for both layers in Fig. \ref{fig_qualit}. The simulated raw point cloud contains missing and wrong points. We observe that SwinDR restores missing parts well, but performs poorly in areas with incorrect data. Our method performs much better in both the first and second layers, yielding a high-quality complete point cloud reconstruction.

\begin{table}[h!t]
\centering
\caption{Depth completion quantitative comparison on DREDS-CatKnown dataset.(all objects/specular\&transparent)}
\label{tab:dep_qua}
\begin{tabular}{c|cccccc}
\hline
Methods & RMSE$\downarrow$ & REL$\downarrow$ & MAE$\downarrow$ \\ \hline
NLSPN&0.010/0.011 &0.009/0.011 &0.006/0.007\\ \hline
LIDF&0.016/0.015& 0.018/0.017& 0.011/0.011 \\ \hline
SwinDR & 0.010/0.010 & 0.008/0.009 & 0.005/0.006\\ \hline
Ours(D1) & \textbf{0.007/0.007} & \textbf{0.006/0.006} & \textbf{0.004/0.004}\\ \hline
\end{tabular}
\vspace*{-10pt}
\end{table}
\begin{table*}[h!t]
\centering
\caption{Evaluation of reconstruction on STD-GraspNet test split (all objects/specular\&transparent).}
\label{tab:dep_qua1}
\begin{tabular}{c|ccc|ccc}
\hline
\multirow{2}{*}{Methods} & \multicolumn{3}{c|}{Evaluation on STD-GraspNet(first-layer)} & \multicolumn{3}{c}{Evaluation on STD-GraspNet(second-layer)}\\ \cline{2-7}
& RMSE$\downarrow$ & REL$\downarrow$ & MAE$\downarrow$ & RMSE$\downarrow$ & REL$\downarrow$ & MAE$\downarrow$ \\ \hline
DREDS& 0.0115/0.0124 & 0.0158/0.0185 & 0.0061/0.0073 &-&-&-\\ \hline
Ours & \textbf{0.0076/0.0080} & \textbf{0.0090/0.0104} & \textbf{0.0036/0.0042} & \textbf{0.0181/0.0175} & \textbf{0.0200/0.0196} & \textbf{0.0090/0.0088} \\ \hline
\end{tabular}
\end{table*}
\subsection{Evaluation Metrics and Baselines for Grasping}
\textbf{Evaluation Metrics.} We evaluate the performance by two metrics: \textbf{Success Rate(SR)}, the ratio of the number of successful grasp and total attempts; and \textbf{Declutter Rate(DR)}, the mean percentage of removed objects across all rounds.

\textbf{Baselines.} 1) \textbf{GSNet}, which takes input single-view raw depth point cloud; 2) \textbf{SwinDR-GSNet} that exploits the previous SOTA depth restoration method for specular and transparent scene, SwinDRNet\cite{dai2022domain}, to first process the raw depth. To be fair, it is fine-tuned on our STD-GraspNet training split to fit GSNet.

\subsection{Simulation Grasping Experiments}
\textbf{Experimental Protocol.} We conduct grasping experiments on STD-GraspNet test-split to evaluate all the methods, which contains 90 scenes divided into 3 categories as seen, similar and novel. Each scene includes 5 to 10 objects with a combination of diffuse, transparent, and specular materials and we uniformly sampled 30 veiwpoints from the total 256 viewpoints for testing. And for each trial in one scene, the robot arm performs grasping and removal of objects until the workspace is cleared, no any grasp detection or two consecutive failures are reached.

\textbf{Results and Analysis.} 
Table \ref{tab:sim} shows our simulation grasping result for different point cloud inputs with GSNet trained with corresponding data. GSNet(RealRaw) with simulated raw depth map which contain incorrect depth or missing on transparent objects, leading the low performance. Although SwinDR improves the depth to some extent, it still suffers from over-smoothing, resulting in a degradation of grasping performance.
For GSNet(SynVisible), input with our predicted visible point cloud achieves similar performance with Oracle(SynVisible), thanks to the original RGB and IR images, learning based stereo matching method and diverse dataset. But we also observe that certain metrics even outperformed the Oracle(SynVisible), which could be caused by the bleeding artifacts near object boundaries, resulting the grasping network to favor grasping pose from the front of objects.  
For GSNet(SynComplete), utilizing our two-layer complete point cloud has the best performance overall. We observe that it outperformed the Oracle(SynVisible). This further shows the potential for improved grasping accuracy with a complete scene representation. However, there is still a noticeable gap compared to the Oracle(SynComplete).
\begin{table*}[h!t]
\centering
\caption{Grasp success rate(\%) of cluster removal in simulation (overall/diffuse/specular/transparent).}
\resizebox{\textwidth}{!}{%
\label{tab:sim}
\begin{tabular}{c|c|ccc|c}
\hline
Depth Source for Testing & Depth Source for Training & Seen & Similar & Novel & $\#$ of grasps \\ \hline
Depth(SimRaw) & GSNet(RealRaw) & 49.7/51.5/52.9/36.7 & 51.7/54.6/51.5/42.6 & 52.3/56.5/52.8/42.1 & 13.8k \\ \hline
SwinDR & GSNet(SimVisible) &  63.9/65.3/68.8/51.9 & 61.8/65.3/64.8/48.1 & 58.5/61.8/62.3/44.7 & 17.7k  \\ \hline
SwinDR & GSNet(RealRaw) &  71.1/72.3/73.9/63.9 & 72.5/74.5/75.3/63.6 &  67.2/68.8/68.6/61.6   & 21.0k \\ \hline
Ours($P_{visible}$) & GSNet(SynVisible) & 91.4/\textbf{91.7}/91.9/90.2 & 90.6/90.2/91.2/90.4 & 87.3/89.8/87.8/82.6   & 22.6k \\ \hline
Ours($P_{comp}$) & GSNet(SynComplete) & \textbf{91.9}/\textbf{91.7}/\textbf{93.0}/\textbf{90.7} & \textbf{91.7}/\textbf{90.9}/\textbf{92.3}/\textbf{92.0} & \textbf{89.6}/\textbf{91.3}/\textbf{89.8}/\textbf{86.9} & 22.4k  \\ \hline
\hline
Oracle(SynVisible) & GSNet(SynVisible) & 91.3/92.1/91.6/89.7 & 89.1/90.3/87.7/89.7 & 88.9/89.9/89.3/86.5  & 22.6k  \\ \hline
Oracle(SynComplete) & GSNet(SynComplete) & 94.1/95.4/93.2/94.1 & 93.1/93.8/92.7/93.5 & 91.7/91.9/93.0/88.7 & 22.5k   \\ \hline
\end{tabular}}
\end{table*}
\begin{table*}[ht]
\centering
\caption{Ablation study on depth completion (all objects/specular\&transparent).}
\resizebox{\textwidth}{!}{%
\label{tab:diff}
\begin{tabular}{c|c|c|c|ccc|ccc}
\hline
\multicolumn{2}{c|}{Input} & \multicolumn{2}{c|}{Training Dataset} & \multicolumn{3}{c|}{GraspNet Dataset Test Split(first-layer)} & \multicolumn{3}{c}{GraspNet Dataset Test Split(second-layer)}\\ \hline
L\&R IR& RGB & DREDS & GraspNet& RMSE$\downarrow$ & REL$\downarrow$ & MAE$\downarrow$  & RMSE$\downarrow$ & REL$\downarrow$ & MAE$\downarrow$\\ \hline
\checkmark & $\times$ & \checkmark & \checkmark  & 0.0084/0.0088 & 0.0097/0.0113 & 0.0038/0.0045 & 0.0187/0.0188 & 0.0206/0.0205 & 0.0092/0.0092 \\ \hline
\checkmark & \checkmark & $\times$ & \checkmark & 0.0081/0.0086 & 0.0101/0.0117 & 0.0040/0.0047  & 0.0190/0.0185 & 0.0222/0.0219 & 0.0099/0.0098  \\ \hline
\checkmark & \checkmark & \checkmark & \checkmark & \textbf{0.0076/0.0080} & \textbf{0.0090/0.0104} & \textbf{0.0036/0.0042} & \textbf{0.0181/0.0175} & \textbf{0.0200/0.0196} & \textbf{0.0090/0.0088} \\ \hline
\end{tabular}}
\end{table*}
\begin{table}[ht]
\centering
\caption{Grasp success rate(\%)/declutter rate(\%) of clutter removal in the real world}
\resizebox{\linewidth}{!}{%
\label{tab:real_grasping}
\begin{tabular}{c|cccc}
\hline
 & Mixed & Transparent & Novel & Overall\\
 & SR(\%)/DR(\%) & SR(\%)/DR(\%) & SR(\%)/DR(\%) & SR(\%)/DR(\%)\\ \hline
GSNet & 50.0/50.0 & 7.7/8.3 & 83.3/35.7 & 39.4/32.5\\ \hline
SwinDR-GSNet & 80.0/92.8 & 41.7/83.3 & 90.0/64.3 & 57.1/80.0 \\ \hline
Ours($P_{1}$) & 87.5/\textbf{100} & 85.71/\textbf{100} & 82.4/\textbf{100} & 85.1/\textbf{100} \\ \hline
Ours($P_{comp}$) & \textbf{93.3}/\textbf{100} & \textbf{100}/\textbf{100} & \textbf{93.3}/\textbf{100} & \textbf{95.2}/\textbf{100} \\ \hline
\end{tabular}}
\vspace*{-20pt}
\end{table}
\subsection{Real Robot Experiments}
To evaluate the performance of our method in real world, we conduct grasping experiments as well. 

\textbf{Experimental Protocol.} In our real-world grasping experiments, there are 18 test objects, including 11 similar objects and 7 novel objects. Novel objects refer to objects not in the training data for both depth restoration and grasp estimation. For the evaluation scenes, we create 6 test scenes, with 2 scenes each of mixed, transparent, and novel objects. The mixed scene contains 2 diffuse objects, 2 specular objects, and 3 transparent objects selected from the group of similar objects. The transparent scene is consisted of 6 similar transparent objects, while the novel scene includes 2 novel specular objects and 5 novel transparent objects. The detailed of test objects and scenes are depicted in Fig. \ref{fig:real_world_setting}. We closely replicate the above 6 scenes for each test baselines. The task is to pick the objects to placement location until the workspace is cleared, no grasp pose generation or 15 attempts are reached.

\textbf{Results and Analysis.} 
Table \ref{tab:real_grasping} shows our real-world grasping evaluation results on the robot, where our method with complete point cloud achieves the highest success rate and declutter rate all test scenes. Compared with original GSNet, our ASGrasp significantly increases the success rate and declutter rate by 55.8\% and 67.5\% in average. In addition, both visible point cloud and complete point cloud as input can achieve 100\% declutter rate. It is worth note that, without depth restoration, the performance of grasping transparent objects is exceedingly low, with a success rate of only 7.7\% and declutter rate of 8.3\% in transparent scene. In contrast, our approach achieves a better performance, achieving over 90\% success rate and declutter rate in transparent scenes. Moreover, it's worth mentioning that SwinDR-GSNet baseline achieves 90\% success rate in novel scene, which is significantly higher than simluation experiments. The reason for this is after several successful grasping, with no further grasp candidate generation, the task ends. In consequence, the novel scene declutter rate of SwinDR-GSNet is lower than its average declutter rate. In addition, our overall success rate in the real world is even higher than in simulation. This is first due to our seamless sim2real domain transfer and partly because of the differences in setups between the real robot and the simulator. The simulator is sensitive that once a collision is detected, the gripper will stop moving. 


\subsection{Ablation Studies}
To analyze the design of our method, we conduct an ablation study on scene reconstruction, considering different inputs and training datasets. 1) To varify the effectiveness of RGB images, we remove the RGB context branch and solely utilize the left and right IR images as input. 2) train the model only on the GraspNet Dataset.

The results are shown in Table \ref{tab:diff}. In comparison to our proposed method, the inclusion of RGB input led to a noticeable enhancement in depth recovery performance at both the first and second layers. Additionally, we observe improvements when incorporating the DREDS dataset, indicating that leveraging diverse data can benefit the learning of scene reconstruction network.

We attempt end-to-end training for the two modules but observe a slight performance drop on the test split of STD-GraspNet. We suspect that this is due to overfitting to the limited size of the finetuning grasping dataset. In this work, we don't further enlarge the grasping training data for a fair comparison with GraspNet-1Billion based method, \textit{e.g.}, GSNet.


\section{CONCLUSIONS}
In this work, we propose an active stereo camera based 6-DoF grasping method, ASGrasp, for transparent and specular objects. We present a two-layer learning based stereo network which reconstructs visible and invisible parts of 3D objects. The following grasping network can leverage rich geometry information to avoid confused grasping. We also propose a large-scale synthetic data to bridge sim-to-real gap. Our method outperforms competing methods on depth metric and clutter removal experiments in both simulator and real world.

\section{ACKNOWLEDGEMENTS}
This work is supported by the National Natural Science Foundation of China (No. 62306016).




\bibliographystyle{ieeetr} 
\balance
\bibliography{bib} %






\end{document}